\def\year{2022}\relax
\documentclass[letterpaper]{article} % DO NOT CHANGE THIS
\usepackage{aaai22}  % DO NOT CHANGE THIS
\usepackage{times}  % DO NOT CHANGE THIS
\usepackage{helvet}  % DO NOT CHANGE THIS
\usepackage{courier}  % DO NOT CHANGE THIS
\usepackage[hyphens]{url}  % DO NOT CHANGE THIS
\usepackage{graphicx} % DO NOT CHANGE THIS
\def\UrlFont{\rm}  % DO NOT CHANGE THIS
\usepackage{natbib}  % DO NOT CHANGE THIS AND DO NOT ADD ANY OPTIONS TO IT
\usepackage{caption} % DO NOT CHANGE THIS AND DO NOT ADD ANY OPTIONS TO IT
\DeclareCaptionStyle{ruled}{labelfont=normalfont,labelsep=colon,strut=off} % DO NOT CHANGE THIS
\usepackage{algorithm}
\usepackage{algorithmic}

\usepackage{newfloat}
\usepackage{listings}
\floatstyle{ruled}
\newfloat{listing}{tb}{lst}{}
\floatname{listing}{Listing}

\usepackage{amsmath}
\usepackage{amssymb}
\usepackage{cleveref}
\usepackage{xspace}
\usepackage{subcaption}
\usepackage{siunitx}
\usepackage{booktabs} 

\newcommand{\EE}{\mathbb{E}}
\newcommand{\NIE}{\textrm{NIE}\xspace}
\newcommand{\doorkey}{\textsc{Doorkey}\xspace}
\newcommand{\twostage}{\textsc{Twostage}\xspace}

\newcommand{\theHalgorithm}{\arabic{algorithm}}

\newtheorem{theorem}{Theorem}%[section]
\renewcommand{\thetheorem}{\arabic{theorem}}

\title{Reinforcement Learning of Causal Variables Using Mediation Analysis}
\author {
	Tue Herlau,\textsuperscript{\rm 1}
	Rasmus Larsen\textsuperscript{\rm 2}
}
\affiliations {
	\textsuperscript{\rm 1} Technical University of Denmark, 2800 Lyngby, Denmark\\
	\textsuperscript{\rm 2} Alexandra Institute, 2300 Copenhagen, Denmark\\
	tuhe@dtu.dk, ralars@dtu.dk
}

\begin{document}
	
	\maketitle
	
	\begin{abstract}
		We consider the problem of acquiring causal representations and concepts in a reinforcement learning setting. 
		Our approach defines a causal variable as being both manipulable by a policy, and able to predict the outcome. 
		We thereby obtain a parsimonious causal graph in which interventions occur at the level of policies.
		The approach avoids defining a generative model of the data, prior pre-processing, or learning the transition kernel of the Markov decision process. 
		Instead, causal variables and policies are determined by maximizing a new optimization target inspired by mediation analysis, which differs from the expected return. 
		The maximization is accomplished using a generalization of Bellman's equation which is shown to converge, and the method finds meaningful causal representations in a simulated environment.  
	\end{abstract}
	
	\section{Introduction} \label{introduction}
	Hard open problems in reinforcement learning, such as distributional shift, generalization from small samples, disentangled representations and counter-factual reasoning, are intrinsically related to causality~\cite{scholkopf2019causality}. Furthermore, causal representations have been emphasized as central to concept acquisition and knowledge representation~\cite{Tenenbaum1279}.
	
	Statistical causal analysis, as developed and popularized by Judea Pearl, assumes that data arises as transformations of noise sources according to a causal graph~\cite{pearl2009causality}. From a practical perspective, describing data generatively as arising from non-linear transformations of i.i.d. noise is an approach that underlies the most successful machine learning models today~\cite{shrestha2019review}. Such an approach has been successfully applied for example in fast concept acquisition~\cite{Tenenbaum1279,lake2015human}, as well as in control~\cite{deisenroth2011pilco, levin}. %Markov decision process (MDP), 
	
	Our approach differs from these in terms of the \emph{scale of modeling}, a term coined by \citet{PetJanSch17}:

	Although traditional examples of causal modeling, such as the $\textsc{smoking}\rightarrow \textsc{tar deposits} \rightarrow \textsc{cancer}$ example~\cite{pearl2018book}, \emph{do} offer a generative process of the few variables included in the analysis, they \emph{do not} offer a generative process of the underlying temporal phenomena (i.e. patient history). The variables are said to be in a \emph{descriptive} relationship to the underlying phenomena, to emphasize that they are not identified as high-level variables in a generative process, but rather features computed from a more complicated underlying phenomena. 
	
	The reduction of the data-generating process to a few abstract primitives in a causal relationship is central to concept acquisition~\cite{Tenenbaum1279}, and more broadly to knowledge representation~\cite{davis1993knowledge}. 
	
	We aim to answer the following question: Can we automatically learn a parsimonious causal model which is descriptive, rather than generative, of the underlying problem, while still capturing relevant causal knowledge?
	
\begin{figure}[t]\centering
		\includegraphics[width=\linewidth]{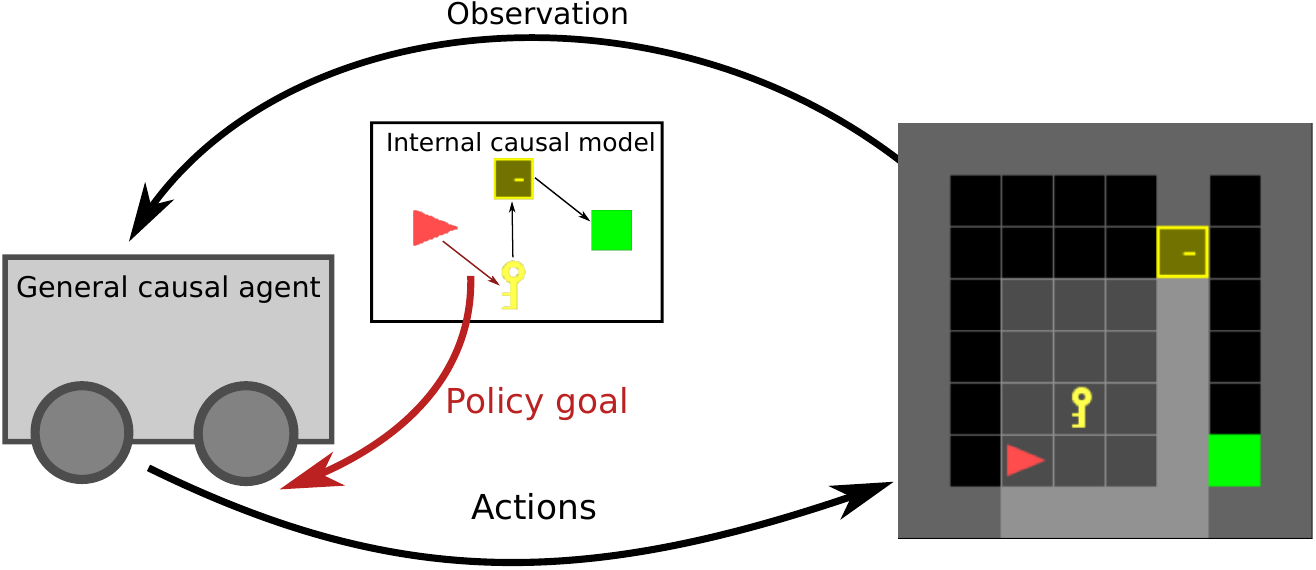}	
		\caption{In the \doorkey environment, the agent (red) must learn to pick up a key to open the door and get to the goal. Our causal agent learns a small, coarse-grained causal network, and uses it when training its policy. }\label{doorkey}
	\end{figure}

	To illustrate, consider the \doorkey environment, \cref{doorkey}. The agent must pick up the key, open the door and go to the goal state in order to receive the reward. Instead of identifying a generative process of the agent moving around the maze, our approach identifies a binary causal variable (for instance, whether the door is opened or not) and builds a small causal graph representing the causal relationship between the identified variable, policy choice and return. The agent is thereby imbued with the causal knowledge that the identified variable is in a causal relationship with the return.

	Our approach\footnote{Code: \url{https://gitlab.compute.dtu.dk/tuhe/causal_nie} } has two central features: First, that we do not identify a causal variable as a latent variable in a generative model of the data, or as a latent factor which arises from maximizing the expected return 
	with respect to the policy. Instead, we replace the expected return with an alternative maximization target, the \emph{natural indirect effect} (\NIE), which is maximized to identify a causal variable. Second, the approach naturally ensures a candidate causal variable represents \emph{a feature of the environment the agent can manipulate}, thereby ensuring the information is relevant for the agent.  
	This distinguishes between \emph{relevant} causal concepts and irrelevant ones. In the \doorkey example (\cref{doorkey}), a variable corresponding to being one step away from the goal would be a necessary cause for completing the environment. However, it would be no easier to manipulate such a variable than simply reaching the goal state. 
	
	To optimize the \NIE in a reinforcement learning setting, we apply suitable generalizations of Bellman's equation. This allows us to apply most actor-critic methods, and specifically, to use an off-policy method based on the $V$-trace estimator~\cite{espeholt2018impala}.

	\paragraph{Related Work:} Determining causal variables has previously been examined in image data from a latent-variable perspective ~\cite{besserve2018counterfactuals,lopez2017discovering} and time-series signals, using (temporal) state aggregation~\cite{zhang2015multi}. However, these approaches apply a latent-variable criteria which is distinct from ours. The problem of determining causal variables has also been investigated from a fairness-perspective, see \citet{zhang2018fairness}.

	In a reinforcement-learning setting, the option-critic architecture considers state-dependent policies similar to ours, but from a non-causal perspective~\citet{bacon2017option}, and \citet{zhang2019learning} learn a state representation using sufficient statistics criteria. Determining latent states to best explain the observations is closely related to the reward machine architecture~\cite{camacho2019ltl,icarte2018using}, which learns binary feature-vector representations in logical, rather than causal, relationships. \citet{Nabi_Kanki_Shpitser_2018} learn policies that optimize a path-specific effect, which is a generalization of the indirect effect. Our approach is different, since we learn both a causal variable and the manipulation policies jointly using a causal criteria.
	
	In recent work, reinforcement learning has been applied for causal discovery in graphs with pre-defined variables, using meta-learning~\cite{dasgupta2019causal} and active learning~\cite{amirinezhad2020active}, for example. \citet{Wang2020ProvablyEC} consider confounded observational data in a reinforcement learning setting, and their approach is noteworthy as they suggest a modified $Q$-learning update. These approaches, however, consider just a handful of variables that can be observed (and manipulated), which is a different setup than the one considered herein.

	\section{Methods} \label{methods}
	Consider a general $\gamma$-discounted episodic Markov decision problem in which states, actions and rewards at time steps $t=0,1,\dots,T$ are denoted $S_t$, $A_t$ and $R_{t+1}$ respectively, and the goal is to maximize the expectation of the return $v_\pi(s) = \EE_{\pi}\left[ G_t \mid S_t=s \right]$ where $G_t = \sum_{k=9}^\infty \gamma^k R_{t+k+1}$. The expectation is with respect to the behavior policy $\pi(a|s) = \textrm{Pr}(A_t=a | S_t=s)$. For easier interpretation, the examples involve sparse reward $+1$, given at the end of the episode in case of successful termination.

	\subsection{Mediation Analysis}
	\begin{figure*}[t]
		\centering
		\includegraphics[width=1\linewidth]{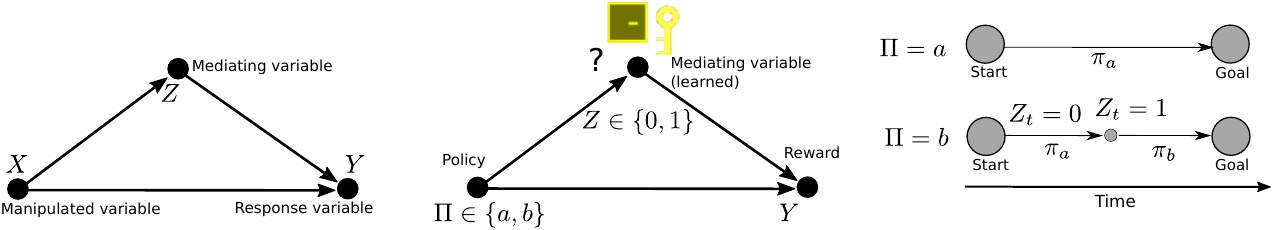}
		\caption{Left: A mediation analysis diagram in which a cause/effect $X \rightarrow Y$ is mediated by a variable $Z$. Center: Application to reinforcement learning: The variable $X$ corresponds to a choice between two policies, the effect $Y$ is the return, and $Z$ is a (learned) variable which influences $Y$. We wish to quantify how the policy choice influences $Y$ through the variable $Z$. Right: $\Pi= a$ indicates that we follow a baseline policy $\pi_a$. This is compared to a policy $\Pi=b$ obtained by following the policy $\pi_b$ until the first time $t$ where $Z_t=1$ occurs, after which we follow $\pi_a$. Both $\pi_a, \pi_b$, and the distribution of $Z_t$, needs to be learned. }\label{diagram1c}
	\end{figure*}
	Mediation analysis~\cite{alwin1975decomposition,pearl2012mediation} deals with decomposing the total causal effect, $p(Y=y|\mathrm{do}(X=x))$, a treatment variable $X$ exerts an outcome variable $Y$ into different causal pathways, which may pass through intermediate \emph{mediating variables}.
	
	In the simplest setting (see \cref{diagram1c}, left), $X$ could be whether a school pupil received extracurricular studies, while $Y$ is their academic performance at the end of the year, and the mediating variable $Z$ could correspond to extra study time.

	Mediation analysis allows us to quantify the extent to which a third variable, such as $Z$, mediates the effect of $X$ on $Y$. The most important measure is the \emph{natural indirect effect}~\cite{pearl2012mediation,pearl2001direct}, which measures the extent to which $X$ influences $Y$, solely through $Z$. For a transition of $X=x$ (starting value) to $X=x'$ (manipulated value), it is defined as expected change in $Y$, affected by holding $X$ constant at its natural value $X=x$, and changing $Z$ to the value it would have attained \emph{had} $X$ been set to $X=x'$. This quantity involves a nested counter-factual, and cannot be estimated in general; however, for specific causal diagrams, it has a closed-form expression. For instance, in the simple case given in \cref{diagram1c} (left), it is defined as~\cite{pearl2001direct}:
	\begin{align}
		\NIE_{x \rightarrow x'}(Y) = \sum_z \EE\left[Y | x,z\right][ P(z | x') - P(z | x)].\label{eqNIE}
	\end{align}
	The \NIE has intuitively appealing properties: It is large when $Z$ is highly influenced by our choice of manipulation $X=x, x'$, meaning that $Z$ is easy to manipulate, and the first term reflects that $Z$ should influence the outcome $Y$. The product implies a trade-off between these two effects. In our application, we let $X=\Pi$ denote our choice of policy, and then use the \NIE to index good (versus bad) choices of the observable variable $Z$ and policies $\Pi$. Hence, we hypothesize that by maximizing the \NIE, rather than the expected return, we can determine relevant causal variables, which correspond to useful concepts for the agent.
	
	\subsection{Causes and Effects in Reinforcement Learning}
	The most natural causal variable to include in a causal diagram is the expected return $Y= G_0$, since manipulating $Y$, and therefore learning which variables are relevant for manipulating $Y$, should remain the eventual goal of the agent. 
	
	Since we consider causal variables as aggregates of many individual states, no single action can reasonably be considered a treatment variable. Rather, we consider a treatment equivalent to the choice to follow policy $\Pi = a$ rather than $\Pi = b$. 
	
	In the following, we focus on the simplest possible case, in which the mediating variable $Z$ is binary,  
	with the meaning that $Z=1$, if the event which $Z$ corresponds to took place during an episode (and otherwise $Z=0$). This is analogous to how \textsc{smoking} is true if the person was smoking \emph{at some point} in the period covered by a study. We therefore define $Z$ as a stopped process
	\begin{align}
		Z = \max\{Z_0, Z_1, \dots Z_T\dots\}.
	\end{align}
	where $Z_t \in \{0,1\}$ for $n=0,1,\dots,T$ denotes 
	whether $Z$ became true at time $t$. $Z_t$ is assumed to only depend on the state, and have distribution $\textrm{Bern}\left( \Phi(s_t) \right)$. %As a consequence, since $Z$ denotes if the given event took place at some point, it holds:
	With these definitions, the causal pathway $\Pi \rightarrow Z \rightarrow Y$ denotes that the choice of policy $\Pi$ influences $Y$ by \emph{obtaining} or \emph{making true} $Z$, whereas $\Pi \rightarrow Y$ means the choice of policy alone influences $Y$, regardless of $Z$. 
	
	\paragraph{Example: }
	Consider the \doorkey environment from \cref{doorkey}. The graphs $\Pi \rightarrow Z \rightarrow Y$ or $\Pi \rightarrow Y$ would reflect either that our choice of $\Pi$ affects the reward $Y$ through $Z$, or that the variable $Z$ is irrelevant, and only the choice of policy matters. The outcome depends on the choice of policy and definition of $Z$. %\emph{What relationship is actually true depends on how we solve the binding problem}, i.e. how we choose to actually define $\Pi$ and $Z$.% Our desiderata for choosing $Z$ is that it behaves somewhat like they key-variable in the \doorkey environment, namely that it stands in non-trivial relationship with both $\Pi$ and $Y$.

	\paragraph{The combined policy:}
	Inspired by the traditional relationship between $X$ and $Z$ in mediation analysis, we assume that if $\Pi = a$ then the agent follows a policy $\pi_a$ which is trained to simply maximize $Y$, and that otherwise, if $\Pi = b$, the agent follows a policy $\pi_b$ which attempts to make $Z$ true (i.e. it is trained with $Z$ as the reward signal). To obtain a well-defined policy for all states, the $\Pi=b$ policy switches back to $\pi_a$ once $Z=1$, see \cref{diagram1c} (right). In other words, we assume that the agent at time step $t$ follows the policy:
	\begin{align}
		\pi = \begin{cases} \pi_a & \mbox{if $\Pi = a$ } \\ \left(1-Z_{0:t} \right)\pi_b +Z_{0:t} \pi_a
			& \mbox{if $\Pi = b$. }\end{cases} \label{eq10}
	\end{align}
	where $Z_{0:t} = \max\{Z_0,\dots,Z_t\}$.
	Since $Z$ and $\Pi$ are binary, the \NIE from \cref{eqNIE} simplifies to~\cite{pearl2001direct}:
	\begin{align}
		\NIE & = \left( \EE\left[Y | Z=1, \pi_a \right] - \EE\left[Y | Z=0, \pi_a \right] \right) \nonumber \\
		&\quad  \times  \left( P(Z=1| \pi_b) -P(Z=1| \pi_a) \right).\label{eq12}
	\end{align}
	Conditioning on $\pi_a$ or $\pi_b$ means that the actions are generated from the given policy. 
	
	The \NIE has the intuitively appealing property of being separated into a product of two simpler terms, which must both be large for the \NIE to be large. The first involves the return, but only conditional on policy $\pi_a$. A high value of the \NIE implies an increased chance of successful completion of the environment, when $Z=1$ relative to $Z=0$. 
	
	The second term involves both policies, but uses $Z$ as a reward signal, which is computed during the episode, and will therefore often be known before the episode is completed. Since this is the only term which involves $\pi_b$, it induces a modular policy, in which $\pi_b$ is trained on a simpler problem.% It furthermore implies the variable $Z$ be more likely to be true under policy $\pi_b$ than $\pi_a$.  % but depending on how $Z$ is defined, may be computed using parts of the trajectory trajectories which are much shorter than  
	
	The \NIE excludes certain trivial definitions of $Z$. For instance, if $Z=Y$ in the \doorkey example, the first term would be maximal. However, in this case, $\pi_a$ and $\pi_b$ would be trained on the same target, and so the second term should be zero. On the other hand, if $Z$ is trained to match states visited by $\pi_b$, which are incidental to the reward, it will not result in a high \NIE, due to the first term. %  accidental ideosyncratic behaviour of l simply reflects a something $\pi_b$ % The target also implies that that $\pi_b$ should be trained to maximize $Z$; i.e. the training of $\pi_b$ in the above is similar to the setup in ordinary reinforcement learning for fixed $Z$. 

	Optimizing the \NIE involves two challenges unfamiliar from traditional reinforcement learning: (i)  %involves some technical challenges which are unfamiliar from expected-return reinforcement learning.
	The first term involves expectations conditional on $Z$. %The Bellman equations (and therefore all value-function based learning) is used to estimate simple \emph{expectation}. Obviously, for the binary case and for fixed $Z$ this could potentially be overcome by dividing the training data into samples where $Z=0$ and $Z=1$, however, this solution is inelegant and does not seem to provide a valid strategy when trying to learn $Z$. 	
	(ii) %	\item 
	The \NIE is optimized both with respect to $Z$ and to $\pi_b$.

	We overcome these by combining two ideas. First, we express the conditional terms using suitable generalizations of Bellman's equation. % easily derived using the standard rules of probability. 
	Secondly, since we optimize policies based on data collected from other policies, we use $V$-trace estimates of the relevant quantities~\cite{espeholt2018impala}.
	
	\subsection{Bellman Updates}
	The value function satisfies the Bellman equation $v_\pi(s) = \EE\left[R_{t+1} + \gamma v_\pi(S_{t+1}) \mid S_t = s\right]$. On comparison with the terms in \cref{eq12}, we see that the \NIE involves conditional expectations. 
	While we could attempt to simply divide the observations according to $Z$ and train two value functions, this method would not provide a way to learn $Z$ itself. To do so, we consider an alternative recursive relationship between the conditional expressions. %a variant of Bellmans equations defined by recursively decomposing the conditional probabilities. 
	
	For times $t \notin \{0,\dots,T\}$ we define $Z_t = 0$. This allows us to introduce the variables %$T$ variables %$Z_k^\infty$
	\begin{align}
		Z_t^\infty = \max\{Z_t,Z_{t+1}, \dots,\} %(1-Z_{k-1}) = \mbox{One of $Z_k,\dots$ is true and $Z_k$ is false}
	\end{align} 
	which are true, provided $Z_{t'}=1$ occurs at a time step following $t$. Note that $Z = Z_0^\infty$.

	Analogous to $v_t$, we define the value functions:
	\begin{align}
		v^{\infty}_t(s_t) & = P(Z^\infty_t =1 | S_t=s_t, Z_{t-1} = 0), \label{eq18} \\
		v^{z}_t(s_t) & = \EE[G_t | S_t=s_t, Z_t^\infty = z, Z_{t-1} = 0].  \label{eq19} 
	\end{align}
	Note that the expressions are conditional on $Z_{t-1} = 0$. The first denotes that the event $Z=1$ will happen in the future given it has not occurred yet, and the second the expected return, given that $Z$ has not happened yet, and either will not $z=0$, or will $z=1$ occur in the future. Note that $v^z_{0}(s_0) = \EE\left[G_0 | Z=z,s_0\right]$ and $v^\infty_0(s_0) = P(Z=1|s_0)$ corresponds to the terms in \cref{eq12}. The value functions satisfy the recursions (see appendix): % which must satisfy: %, and a little algebra show they satisfy:
	\begin{subequations} \label{eqx19}
		\begin{align}%{2}
			v^\infty_t(s_t) & = \Phi(s_t) + \bar \Phi(s_t) \EE\left[ v^\infty_{t+1}(S_{t+1} )| s_t\right],\label{eq20a}   \\ 
			v^{1}_t(s_t) & =  \frac{ V(s_t) \Phi(s_t) }{V_t^\infty(s_t) }   \label{eq21a}  \\
			+ \frac{  1\!-\! \Phi(s_t)  }{V_t^\infty(s_t) } & \EE\left[ v^\infty_{t+1}(S_{t+1}) \left( R_{t+1} + \gamma v^{1}_{t+1}(S_{t+1} ) \right) \mid s_t   \right], \nonumber \\
			v^{0}_t(s_t) & = 
			\frac{ 1- \Phi(s_t)  }{ 1\!-\!v_t^\infty(s_t) } \label{eq21b} \\
			\times \EE[
			\ ( 1-&v_{t+1}^\infty(S_{t+1}) )( 
			R_{t+1}  + \gamma v^{0}_{t+1}(S_{t+1}) ) \mid s_t ].  \nonumber
		\end{align}
	\end{subequations}
	
	The new recursions have the same structure as Bellman's equation, but contain mutually dependent terms. If $v_t^\infty$ and $v_t$ were exactly estimated, the iterative policy evaluation methods corresponding to \cref{eq21a,eq21b} would easily be found to be contractions with constant $\gamma$, but the updates also converge when $v^z$, $v^\infty$ and $v$ are all bootstrapped. A proof can be found in the appendix. % \sref{Athm2} 
	
	\begin{theorem}[Convergence, informal]	\label{T1}
		Assuming $\gamma<1$ and $0<\Phi<1$, all states/actions are visited infinitely often, and $v_\pi, v^\infty_\pi, v^z_\pi$ in \cref{eqx19} are all replaced by randomly initialized bootstrap estimates. Then, (i) the operators \cref{eq21a,eq21b} converge at a geometric rate to the true values $v^z_\pi$, and (ii) the corresponding online method obtained by replacing the expectations with sample estimates, converges to the true values, provided the learning rates satisfy Robbins-Monro conditions. 
	\end{theorem}

	\subsection{Off-Policy Learning Using $V$-Trace Estimators}
	The overall approach is to learn neural approximations of $v^\infty$, $v$ and $v^z$, as defined in \cref{eqx19}. This is most easily done by observing that the Bellman-like recursions in \cref{eqx19} all have the form:
	\begin{align}
		v_t(s_t) & =\EE_\mu\left[H_t(s_t,S_{n+1})  + G_t(s_t, S_{t+1}) v_{t+1}(S_{t+1} )  )  \middle| s_t\right] \label{eq22a}
	\end{align}
	where actions are generated using a behavioral policy $\mu$. %Since we will discuss off-policy learning momentarily, we assumed all actions are generated using a \emph{behavior policy} $\mu$. %Suppose actions in 
	Expanding the right-hand side $n$ times, allows us to define the $n$-step return~\cite{espeholt2018impala}:
	\begin{align}  
		v_t(s_t) & =\EE\left[ \sum_{i=t}^{t+n-1} H_i \prod_{\ell=t}^{i-1} G_\ell + v_{t+n}(S_{t+n})  \prod_{\ell = t}^{ t+n-1} G_\ell   \middle| s_k\right], \label{eq35}
	\end{align}
	which reduces to \cref{eq22a} if $n=1$. Supposing the current target policy is $\pi$, experience is collected from the behavior policy $\mu$, and then \cref{eq35} can be used as an estimate of the return, corresponding to $\pi$, by using importance sampling. To reduce variance, we use a $V$-trace type estimator, inspired by~\citet{espeholt2018impala}: 
	\begin{subequations}
		\begin{align}
			V_t(s_t) &= v(s_t) + \sum_{i=t}^{t+n-1} \left( \prod_{\ell=t}^{i-1} c_\ell G_{\ell} \right)	\delta_i \label{eq24a} \\
			\delta_i & = \rho_i \left[ H_i(s_i, s_{i+1}) + G_i v(S_{i+1})- v(S_i) \right]  
		\end{align} \label{eq22}
	\end{subequations}
	where $c_\ell$ and $\rho_k$ are truncated importance sampling weights: 
	\begin{align} 
		\rho_t = \min\left\{\bar \rho, \frac{\pi(a_t | s_t)  }{\mu(a_t | s_t) }  \right\},\
		c_t = \min\left\{\bar c, \frac{\pi(a_t | s_t)  }{\mu(a_t | s_t) }  \right\}, \nonumber
	\end{align}
	and $\bar \rho \geq \bar c$ are parameters of the method. In the on-policy case, where $\mu = \pi$, the $V$-trace estimate \cref{eq24a} reduces to $\sum_{i=t}^{t+n-1} H_i \prod_{\ell=t}^{i-1} G_\ell + v_{t+n}  \prod_{\ell = t}^{ t+n-1} G_\ell$,  
	and is therefore a direct estimate of \cref{eq22a}. In the general case, the method provides a biased estimate, when $\bar \rho, \bar c < \infty$, but analogous to \citet{espeholt2018impala}, the stationary value function can be analytically related to the true value function. The result is summarized as: (see the appendix for further details) 
	\begin{algorithm}[t] 
		\caption{Causal learner}\label{alg1}
		\begin{algorithmic}[1]
			\STATE Initialize policy networks $\pi_a$ and $\pi_b$ (and corresponding critic networks)
			\STATE Initialize networks $v$, $v^{0}$, $v^{1}$, $v^{\infty}$ to estimate $v_\pi$, $v^{z}_\pi$,  and $V^\infty_\pi$		
			\STATE Initialize causal variable network $\Phi$ 
			\REPEAT
			\STATE Collect experience from $\pi_a$ and add to replay buffer
			\STATE Sample experience from replay buffer $\tau$
			\STATE Train $\pi_a$ (and critic) using AC2
			\STATE Calculate reward signal for $\pi_b$ from $\tau$ using \cref{eq41} and train $\pi_b$ (and critic)
			\STATE Train $v$, $v^{z}$, $v^{\infty}$ using $n$-step $V$-trace estimates \cref{eq24a}, computed using \cref{eq27}, using definitions of $H_t$ and $G_t$ implied by \cref{eqx19} and experience $\tau$
			\STATE Train parameters in causal variable network $\Phi$ by maximizing \cref{eq12}, where each term has been replaced by the respective $V$-trace estimate computed using \cref{eqx19,eq24a}
			\UNTIL{forever}
		\end{algorithmic}
	\end{algorithm}
	
	\begin{theorem}[$V$-trace convergence, informal]	\label{T2}
		Assume that experience is generated by a behavior policy $\mu$, that  $\gamma<1$, $0<\Phi<1$, all states/actions are visited infinitely often, and that $v_\pi, v^\infty_\pi, v^z_\pi$ in \cref{eqx19} are all replaced by randomly initialized bootstrap estimates. Then, if we apply \cref{eq24a} iteratively~\footnote{$x \leftarrow_\alpha y$ is equivalent to $x = x(1-\alpha) + \alpha y$} on the bootstrap estimate of $V^z$ 
		\begin{align}
			V^z(s) \leftarrow_\alpha V^z(s) + \sum_{t=0}^{\infty} \prod_{\ell =0}^{t-1} c_\ell G_\ell^z \delta_\ell,
		\end{align}
		where $H_\ell^z$ and $G^z_\ell$ are computed using $V$-trace estimates of $v_\pi$ and $v^\infty_\pi$, it implies that $V^z$ converges to a biased estimate of $v^z_\pi$, and if $\bar \rho, \bar c \rightarrow \infty$, then $V^z \rightarrow v_\pi^z$. 
	\end{theorem}
	
	To practically compute the $V$-trace estimates, we start from $T$ and proceed to $t$: %starting from $n+N$ and proceeding to $n$, as
	\begin{align}
		V_t = v_t + \delta_t + G_t c_t (V_{t+1} - v_{t+1}). \label{eq27}
	\end{align}

	\subsection{Combined Method} \label{s25}
	The policy $\pi_b$ in \cref{eq10} is trained in an episodic environment to maximize $Z$. Since the variable $Z$ is multiplicative over individual time steps, we train $\pi_b$ by decomposing the multiplicative cost using a stick-breaking construction: 
	\begin{align}
		r_{t+1}^b = \Phi(s_t) \prod_{k=0}^{t-1} (1-\Phi(s_k)), \label{eq41}
	\end{align}
	which satisfies $\sum_{t=0}^\infty r_{t+1} = P(Z=1|\tau)$. Training on this reward signal means that the policy $\pi_b$ will attempt to maximize the term $P(Z=1|\pi_b) - P(Z=1|\pi_a)$ in \cref{eq12}. We can therefore train both $\pi_a$ and $\pi_b$ with an actor-critic method, using their respective reward signals, whereby the critics estimate of the return are trained against the $V$-trace estimate, as computed using \cref{eq27}.

	To maximize the \NIE with respect to $\Phi$, we introduce networks $v$, $v^z$ and $v^\infty$, to approximate $v_\pi$, $v^\infty_\pi$ and $v^z_\pi$. These are trained using ordinary gradient descent against their $V$-trace targets, computed by \cref{eq27}. The same $V$-trace estimates can be used to re-write the \NIE in \cref{eq12}, to an expression which directly depends on $\Phi$, and can therefore be trained using stochastic gradient descent.  
	For instance, $\EE[Y|Z=1,\Pi=a,s_0] = v_{\pi_a}^1(s_0)$ is equivalent to $V^{z=1}_t(s_0)$, computed using \cref{eq27}, and the definitions of $H_t$ and $G_t$ implied by \cref{eq21a}, and $\EE\left[Z=1|\Pi= a\right]$ can be replaced by $V_0^\infty$, computed using \cref{eq20a}. The pseudo-code of the method can be found in \cref{alg1}.
	Note that to prevent premature convergence, and speed up convergence when both factors in the \NIE are small, we train on a surrogate cost function which includes entropy terms for $\Phi$, $\pi_a$ and $\pi_b$. Full details can be found in the appendix.

	\section{Experiments}\label{experiments}
	
	\begin{figure*}
		\centering
	\end{figure*}

	\begin{figure*}
		\centering
\includegraphics[width=.96\linewidth]{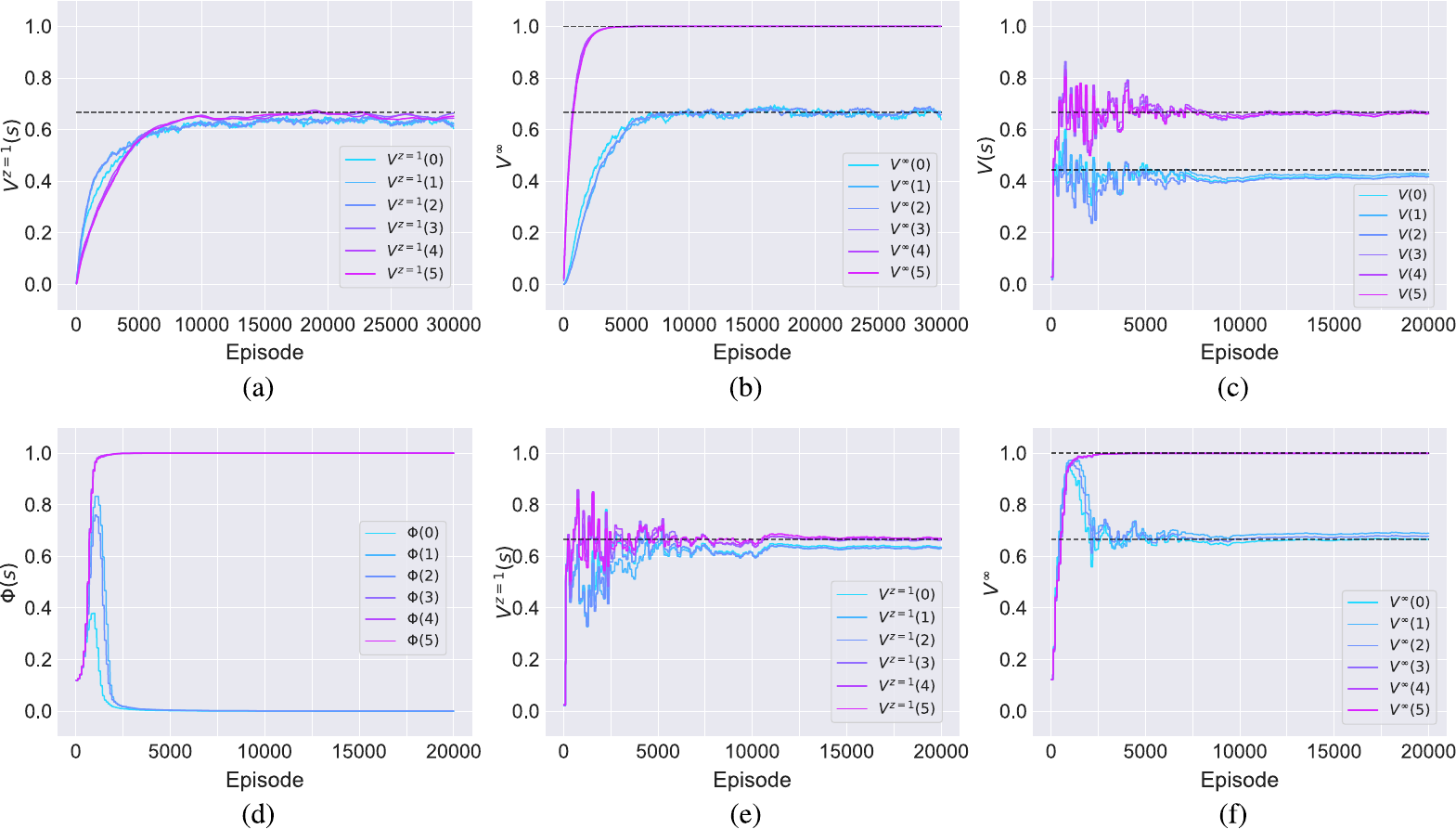}
\begin{subfigure}{0\textwidth}\refstepcounter{subfigure}\label{fig:twostage_a}\end{subfigure}%
\begin{subfigure}{0\textwidth}\refstepcounter{subfigure}\label{fig:twostage_b}\end{subfigure}%
\begin{subfigure}{0\textwidth}\refstepcounter{subfigure}\label{fig:twostage_c}\end{subfigure}%
\begin{subfigure}{0\textwidth}\refstepcounter{subfigure}\label{fig:twostage_d}\end{subfigure}%
\begin{subfigure}{0\textwidth}\refstepcounter{subfigure}\label{fig:twostage_e}\end{subfigure}%
\begin{subfigure}{0\textwidth}\refstepcounter{subfigure}\label{fig:twostage_f}\end{subfigure}%
		\caption{ (a-b) Trace plots of $v^{1}$ and $v^{\infty}$ for the tabular \twostage environment obtained using \cref{eq35}, with a given $\Phi$. (c-f) Estimates with neural function approximators for the value functions in the \twostage environment, while $\Phi$ is being learned.
		}
	\end{figure*}
	
	We test the value function recursions in \cref{eqx19} on a simple Markov reward process dubbed \twostage corresponding to an idealized version of the \doorkey environment. In \twostage, the states are divided into two sets $S_A$ and $S_B$. The initial state is always in $S_A$, and the environment can either transition within sets ($S_A \rightarrow S_A$, $S_B \rightarrow S_B$) with a fixed probability, or from set $S_A$ to $S_B$, with a fixed probability. From $S_B$, there is a chance to terminate successfully with a reward of $+1$, and from all states there is a chance to terminate unsuccessfully with a reward of $0$. 
	
	The transition from states in $S_A$ to $S_B$, creates a bottleneck distinguishing successful and unsuccessful episodes, much like unlocking the door in the \doorkey environment. The transition probabilities are chosen such that $p(R=1| s \in S_B) = p(s \in S_B | s \in S_A)  = \frac{2}{3}$ and $p(R=1 | s \in S_A) = \frac{4}{9}$, see the appendix for further details.

	\subsection{Tabular Learning} \label{sec32}% \iftoggle{arxiv}{}{  (see \texttt{tabular\_agent\_td.py})} }
	As a first example, we will consider simple estimation of the conditional expectations, using the Bellman recursions. We condition on whether the state enters $S_B$ at a later time, i.e. $\EE[Y| \mbox{$s_t \in S_B$ for some $t > 0$}, S_0 = s_0]$, which is equivalent to $v^1(s_0)$, since we define $\Phi(s) = 1_{S_B}(s)$. In this case, the Bellman updates from \cref{eqx19}) for a transition $S_t = s$ to $S_{t+1}=s'$, $R_{t+1} = r$ are
	\begin{align}
		V(s) & \underset{\alpha }{\leftarrow } r + \gamma V(s') \nonumber  \\% \label{eq35a} \\
		V^\infty(s) & \underset{\alpha}{\leftarrow } \Phi(s)  + (1-\Phi(s) ) V^\infty(s') \label{eq35b} \\ 
		\!\!V^{1}(s) & \underset{\alpha}{\leftarrow }  
		\frac{ 
			V(s) \Phi(s)\! +\! (1\! -\! \Phi(s) )V^\infty(s_{t} ) \left(  r\! +\!  \gamma V^{1}(s' )  \right) }{ V^\infty(s) } \nonumber %\label{eq35c}
	\end{align}
	As anticipated by \cref{T1}, iterating these updates, the value functions converge to their analytically expected values, as can be seen in \cref{fig:twostage_a,fig:twostage_b}, in which we plot $v^1$ and $v^\infty$. The dashed lines represent the true (analytical) values, and the different colored lines represent the different states. 
	In the case of $v^1$, the expectation estimated is the probability of successful completion, given that we begin in any state and \emph{at some point} enter $S_B$; in other words, the information we condition on is something which only occurs \emph{at a later point} in the episode, from the perspective of an observation $s\in S_A$, and therefore the correct estimation of this probability is not simply a matter of computing the return for a state starting in $S_B$. %The result is shown in \cref{fig5}, and shows good convergence for all states to the theoretically predicted value indicated by the horizontal lines. 

	\subsection{Learning $\Phi$ Using $V$-Trace Estimation} \label{sec33} % \iftoggle{arxiv}{}{ (see \texttt{run\_twostage\_learn\_phi.py}) } } 

	The second example extends the \twostage example to also learn the causal variable $\Phi$ using \cref{alg1}. Since the environment is a MRP, we discard terms involving $\pi_b$, and the objective $\Delta_Y$ therefore becomes: %and only include the non-zero term of the objective 
	\begin{align}
		\EE\left[Y|Z\!=\!1 \right]\! -\! \EE\left[Y|Z\!=\!0\right]\! =\! \EE_{s_0} \left[ V^{1}(s_0)\! -\! V^{0}(s_0) \right]. \label{eq45}
	\end{align}
	The expectation is unrolled, using the $V$-trace approximation, and directly optimized with respect to the parameters $(w_s)_s$ of $\Phi$, using the parameterization $\Phi(s) = \frac{1}{1+\exp(-w_s)}$.

	The value function approximation is quickly learned (see \cref{fig:twostage_c}), showing convergence to the analytical values.
	The quantities $V^\infty$ and $V^{z}$ both depend on $\Phi$, and will therefore only begin to converge after $\Phi$ begins to converge (see \cref{fig:twostage_d}). Since the conditional expectations $V^z$ depend on $V^\infty$, they will converge relatively slower, but both will eventually converge to their expected value when the learning rate is annealed, see \cref{fig:twostage_e}.

	\subsection{Causal Learning and the \doorkey Environment}
	To apply \cref{alg1} to the \doorkey environment, we first have to parameterize the states. The environment has $|\mathcal A| = 5$ actions, and we consider a fully-observed variant of the environment. We choose the simplest possible encoding, in which each tile, depending on its state, is one-hot encoded as an 11-dimensional vector. This means that an $n\times n$ environment is encoded as an $n\times n \times 11$-dimensional sparse tensor, and we include a single one-hot encoded feature to account for the player orientation. Further details can be found in the appendix. Episode length is 60 steps.

	Since the environment encodes orientation, player position and goal position separately, and since specific actions must be used when picking up the key and opening the door, the environment is surprisingly difficult to explore and generalize in. We train an agent using A2C \citep{mnih2016asynchronous} with 1-hidden-layer fully connected neural networks, which results in a completion rate of about $0.25$ within the episode limit. We also attempted to train an agent using the Option-Critic framework \citep{bacon2017option}, a relevant comparison to our method, but failed to learn options which solved the environment better than chance.

	After an initial training of $\pi_a$, we train $\Phi$ and $\pi_b$ by maximizing the \NIE, using \cref{alg1}. Training parameters can be found in the supplementary material. To obtain a fair evaluation on separate test data, we simulate the method on $200$ random instances of the \doorkey  environment, and use Monte-Carlo roll-outs of the policies $\pi_a$ and $\pi_b$ to estimate the quantities $\EE \left[Z=1 \mid \Pi = a \right]$, $\EE \left[Z=1 \mid \Pi = b \right]$. This allows us to estimate the \NIE on a separate test set. 
	
	To examine whether the obtained definition of $Z$ is non-trivial, we compare it against a natural alternative that learns $Z$ by maximizing the cross-entropy of $Z$ and $Y$, % $P(Y|Z=1)$ and $P(Y|Z=0)$. 
	\begin{align}
		-\EE_\tau\left[Y(\tau) \log P(Z=z|\tau)\right]. \label{eq46}
	\end{align}
	Since $Y$ is binary, this corresponds to determining $\Phi$ as the binary classifier which 
	separates successful ($Y=1$) episodes from unsuccessful episodes ($Y=0$), i.e. ensures that the first factor of the \NIE \cref{eq45} is large. 
	
	\begin{table*}[t]
		\centering
\begin{tabular}{ lcccccc }
	\toprule Method
	&  $\mathbb{E}[Y\mid Z=1]$
	&  $\mathbb{E}[Y\mid Z=0]$
	&  $\Delta_Y$
	&  $\mathbb{E}_{\pi_a}[Z]$
	&  $\mathbb{E}_{\pi_b}[Z]$
	&  NIE
	\\ \midrule Causal Learner
	&  \SI{0.410\pm 0.030}{}
	&  \SI{0.000\pm 0.000}{}
	&  \SI{0.410\pm 0.030}{}
	&  \SI{0.550\pm 0.020}{}
	&  \SI{0.790\pm 0.020}{}
	&  \SI{0.098\pm 0.010}{}
	\\  Cross-entropy
	&  \SI{0.560\pm 0.080}{}
	&  \SI{0.130\pm 0.030}{}
	&  \SI{0.430\pm 0.100}{}
	&  \SI{0.270\pm 0.040}{}
	&  \SI{0.270\pm 0.030}{}
	&  \SI{0.011\pm 0.008}{}
	\\ \bottomrule\end{tabular}
		\caption{Performance of causal agent on the \doorkey environment and standard deviation of the mean.}\label{tbl_12}
	\end{table*}
	
	The results of both methods can be found in \cref{tbl_12} (results averaged over 10 restarts with different seeds). The causal learner obtains a value of the \NIE that is significantly different from zero in all runs. While the absolute value is small, this can be attributed to the \NIE being a product of two factors which are both small. Considering the first two terms, we observe that the causal variable $Z=1$ is a necessary condition for completing the environment, while the corresponding variable for the cross-entropy target can be false, yet the agent is still able to successfully complete the environment. 
	
	We also notice that the cross-entropy based learner outperforms the causal target, in terms of obtaining a proper separation between good versus bad trajectories, i.e. a higher value of $\Delta_Y$. % = \EE_\tau\left[Y| Z=1\right] - Y(\tau) \log P(Z=z|\tau)\right]
	This is expected, since cross-entropy is an efficient cost-function for a binary classification problem. 
	
	However, the causal variable $Z$, which is learned by the cross-entropy learner, does not present a suitable target for the policy $\pi_b$. Indeed, the variable $Z$ becomes true at the same rate under $\pi_a$ and $\pi_b$ (all policies are trained using the same settings). This can be accounted for by recalling that the environment is random, and that the variable $Z$ learned by the causal learner represents a relatively stable feature of the environment (such as picking up the key, opening the door, etc.), whereas the cross-entropy trained variable $Z$ corresponds to a combination of features in the environment which presents a less suitable optimization target.

\begin{figure}[t]
		\includegraphics[width=.95\linewidth]{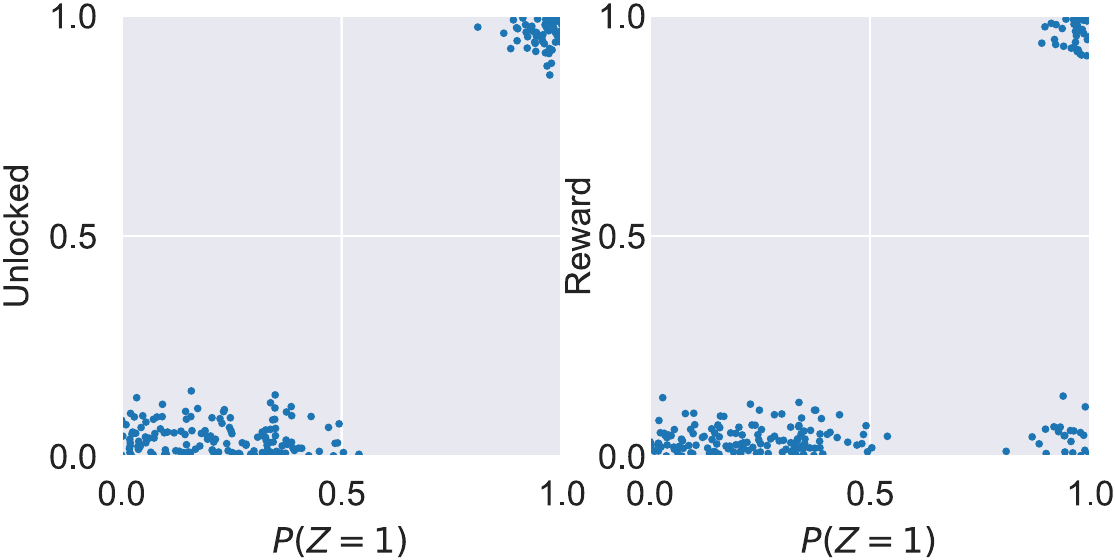}
		\caption{Scatter plot of $P(Z=1)$ and (left) chance of unlocking the door, and (right) chance of successfully reaching the goal state. The causal variable $Z$  appears to correspond to opening the door. }\label{fig11} % also b, c. 
	\end{figure}

	To obtain insight in the causal variable we learn, we plot $P(Z=1)$ both against the reward, and whether the door was opened in this particular run (jitter added for easier visualization). The results can be found in \cref{fig11}. As indicated, the learned causal variable correlates well with whether the door is opened or not, and not as well with the total reward. In other words, the method is able to learn that the feature of whether the agent has opened the door acts as a mediating cause in terms of completing the environment. This is a natural result, considering this task is necessary in order for the agent to complete the environment.

	The fact that the causal variable corresponds to a meaningful objective, is reinforced by examining the total reward obtained from either following policy $\Pi =a$, or the joint policy $\Pi=b$ (see \cref{tbl_13}). Although the difference is slight, we observe a small increase in accumulated reward for the joint policy.

	\begin{table}
		\centering
\begin{tabular}{ lcc }
	\toprule Method
	&  $\mathbb{E}[Y| \Pi=a ]$
	&  $\mathbb{E}[Y| \Pi=b ]$
	\\ \midrule Causal Learner
	&  \SI{0.240\pm 0.020}{}
	&  \SI{0.300\pm 0.030}{}
	\\  Cross-entropy
	&  \SI{0.230\pm 0.020}{}
	&  \SI{0.240\pm 0.010}{}
	\\ \bottomrule\end{tabular}
		\caption{Reward obtained in the \doorkey environment.}\label{tbl_13}
	\end{table}
	
	\section{Conclusion} \label{conclusion}
	Since all causal conclusion depends on assumptions that are not testable in observational studies~\cite{pearl2009causal}, it is natural to ask why we are justified in believing that a particular method finds a \emph{causal} representation of the environment. 
	
	In work involving pre-defined variables, such justification can be found either through external distributional assumptions about the relationship between the structure of the model and the data it generates~\cite{spirtes2000causation}, or because the model belongs to a class of models of which so many examples have been observed, that meta-learning allows the structure to be identified~\cite{dasgupta2019causal}.

	In contrast, our work assumes a specific causal diagram which, along with the definition of $Z$ and $Y$, ensures that $Z$ is imbued with a natural interpretation as the mediating causal factor of the causal pathway from $X$ to $Y$.
	
	A more fundamental question is \emph{why} a parsimonious model of causal knowledge, such as the $\textsc{smoking}/\textsc{cancer}$ example, is preferable to a detailed causal model of patient history. Indeed, if we adopt the view that a model should best fit the MDP (i.e. a generative view), it is difficult to see why parsimony would be preferred. 
	
	Although we do not claim to have a definite answer, our approach differentiates situations in which it can find causal knowledge from those in which it cannot, \emph{without} referencing a generative/best fit criteria. Specifically, for a variable $Z$ to be identified, it must be so relative to a policy $\pi_a$, as \emph{something the agent could potentially do}, and is associated with a high reward ($\EE[Y|Z=1] > \EE[Y|Z=0]$), but it \emph{might not do it under its baseline behavior $\pi_a$}. As a consequence, the policy $\pi_a$ must be sub-optimal in order for the agent to determine a causal model.

	At a glance, this may seem like a flaw in the method, but the idea that causation is ill-defined when one has a precise description of the physical world is old~\cite{russell1913notion}, and closely matches our common sense: When a child learns to ride a bicycle, a potential causal explanation for a fall, such as \emph{steering too far from the center of the lane} ($Z=0$) and hitting the curb ($Y=0$), is only relevant in case the child \emph{could have} taken \emph{better} actions to keep near the center of the lane ($Z=1$). To put this in a different way, if the agent knows enough about the environment to have an optimal policy, a coarse-grained causal model cannot offer the agent any benefits because there are no policy decisions to improve. 
	
	This example hopefully clarifies a point made during review, namely how a choice of policy, $\Pi = \pi_a$ or $\Pi = \pi_b$ can act as a cause: The policy itself is not a cause, but rather the binary variable which denotes which policy is followed is treated as a cause.

	We have argued that the \NIE offers a novel way to define coarse-grained causal knowledge. In identifying a causal variable, our method learns a policy to manipulate it, and the variables are learned directly from experience \emph{without} requiring specification of a generative process. We have shown the conditional expectations involved can be estimated using an $n$-step temporal difference methods, and that the method has convergence properties  comparable to TD learning (but with worse constants). We found that the method was able to learn a causal variable which was both sensible and relevant for solving a task, and did so better than a natural (non-causal) alternative method. An independent experiment on a test-set indicated the causal representation is associated with an increased \NIE, and that the causal representation is relevant for the task in that it resulted in a performance increase. %

	\bibliography{references}

\begin{thebibliography}{30}
\providecommand{\natexlab}[1]{#1}

\bibitem[{Alwin and Hauser(1975)}]{alwin1975decomposition}
Alwin, D.~F.; and Hauser, R.~M. 1975.
\newblock The decomposition of effects in path analysis.
\newblock \emph{American sociological review}, 37--47.

\bibitem[{Amirinezhad, Salehkaleybar, and
  Hashemi(2020)}]{amirinezhad2020active}
Amirinezhad, A.; Salehkaleybar, S.; and Hashemi, M. 2020.
\newblock Active Learning of Causal Structures with Deep Reinforcement
  Learning.
\newblock \emph{arXiv preprint arXiv:2009.03009}.

\bibitem[{Bacon, Harb, and Precup(2017)}]{bacon2017option}
Bacon, P.-L.; Harb, J.; and Precup, D. 2017.
\newblock The option-critic architecture.
\newblock In \emph{Proceedings of the AAAI Conference on Artificial
  Intelligence}, volume~31.

\bibitem[{Besserve et~al.(2020)Besserve, Mehrjou, Sun, and
  Sch{\"o}lkopf}]{besserve2018counterfactuals}
Besserve, M.; Mehrjou, A.; Sun, R.; and Sch{\"o}lkopf, B. 2020.
\newblock Counterfactuals uncover the modular structure of deep generative
  models.
\newblock In \emph{Eighth International Conference on Learning Representations
  (ICLR 2020)}.

\bibitem[{Camacho et~al.(2019)Camacho, Icarte, Klassen, Valenzano, and
  McIlraith}]{camacho2019ltl}
Camacho, A.; Icarte, R.~T.; Klassen, T.~Q.; Valenzano, R.~A.; and McIlraith,
  S.~A. 2019.
\newblock LTL and Beyond: Formal Languages for Reward Function Specification in
  Reinforcement Learning.
\newblock In \emph{IJCAI}, volume~19, 6065--6073.

\bibitem[{Dasgupta et~al.(2019)Dasgupta, Wang, Chiappa, Mitrovic, Ortega,
  Raposo, Hughes, Battaglia, Botvinick, and Kurth-Nelson}]{dasgupta2019causal}
Dasgupta, I.; Wang, J.; Chiappa, S.; Mitrovic, J.; Ortega, P.; Raposo, D.;
  Hughes, E.; Battaglia, P.; Botvinick, M.; and Kurth-Nelson, Z. 2019.
\newblock Causal reasoning from meta-reinforcement learning.
\newblock \emph{arXiv preprint arXiv:1901.08162}.

\bibitem[{Davis, Shrobe, and Szolovits(1993)}]{davis1993knowledge}
Davis, R.; Shrobe, H.; and Szolovits, P. 1993.
\newblock What is a knowledge representation?
\newblock \emph{AI magazine}, 14(1): 17--17.

\bibitem[{Deisenroth and Rasmussen(2011)}]{deisenroth2011pilco}
Deisenroth, M.; and Rasmussen, C.~E. 2011.
\newblock PILCO: A model-based and data-efficient approach to policy search.
\newblock In \emph{Proceedings of the 28th International Conference on machine
  learning (ICML-11)}, 465--472. Citeseer.

\bibitem[{Espeholt et~al.(2018)Espeholt, Soyer, Munos, Simonyan, Mnih, Ward,
  Doron, Firoiu, Harley, Dunning et~al.}]{espeholt2018impala}
Espeholt, L.; Soyer, H.; Munos, R.; Simonyan, K.; Mnih, V.; Ward, T.; Doron,
  Y.; Firoiu, V.; Harley, T.; Dunning, I.; et~al. 2018.
\newblock Impala: Scalable distributed deep-rl with importance weighted
  actor-learner architectures.
\newblock In \emph{International Conference on Machine Learning}, 1407--1416.
  PMLR.

\bibitem[{Icarte et~al.(2018)Icarte, Klassen, Valenzano, and
  McIlraith}]{icarte2018using}
Icarte, R.~T.; Klassen, T.; Valenzano, R.; and McIlraith, S. 2018.
\newblock Using reward machines for high-level task specification and
  decomposition in reinforcement learning.
\newblock In \emph{International Conference on Machine Learning}, 2107--2116.
  PMLR.

\bibitem[{Lake, Salakhutdinov, and Tenenbaum(2015)}]{lake2015human}
Lake, B.~M.; Salakhutdinov, R.; and Tenenbaum, J.~B. 2015.
\newblock Human-level concept learning through probabilistic program induction.
\newblock \emph{Science}, 350(6266): 1332--1338.

\bibitem[{Levine et~al.(2016)Levine, Finn, Darrell, and Abbeel}]{levin}
Levine, S.; Finn, C.; Darrell, T.; and Abbeel, P. 2016.
\newblock End-to-End Training of Deep Visuomotor Policies.
\newblock \emph{J. Mach. Learn. Res.}, 17(1): 1334–1373.

\bibitem[{Lopez-Paz et~al.(2017)Lopez-Paz, Nishihara, Chintala, Scholkopf, and
  Bottou}]{lopez2017discovering}
Lopez-Paz, D.; Nishihara, R.; Chintala, S.; Scholkopf, B.; and Bottou, L. 2017.
\newblock Discovering causal signals in images.
\newblock In \emph{Proceedings of the IEEE Conference on Computer Vision and
  Pattern Recognition}, 6979--6987.

\bibitem[{Mnih et~al.(2016)Mnih, Badia, Mirza, Graves, Lillicrap, Harley,
  Silver, and Kavukcuoglu}]{mnih2016asynchronous}
Mnih, V.; Badia, A.~P.; Mirza, M.; Graves, A.; Lillicrap, T.; Harley, T.;
  Silver, D.; and Kavukcuoglu, K. 2016.
\newblock Asynchronous methods for deep reinforcement learning.
\newblock In \emph{International conference on machine learning}, 1928--1937.
  PMLR.

\bibitem[{Nabi, Kanki, and Shpitser(2018)}]{Nabi_Kanki_Shpitser_2018}
Nabi, R.; Kanki, P.; and Shpitser, I. 2018.
\newblock Estimation of Personalized Effects Associated With Causal Pathways.
\newblock \emph{Uncertainty in artificial intelligence: proceedings of the...
  conference. Conference on Uncertainty in Artificial Intelligence}, 2018.

\bibitem[{Pearl(2001)}]{pearl2001direct}
Pearl, J. 2001.
\newblock Direct and indirect effects.
\newblock In \emph{Proceedings of the Seventeenth conference on Uncertainty in
  artificial intelligence}, 411--420.

\bibitem[{Pearl(2009)}]{pearl2009causality}
Pearl, J. 2009.
\newblock \emph{Causality: Models, Reasoning and Inference}.
\newblock USA: Cambridge University Press, 2nd edition.
\newblock ISBN 052189560X.

\bibitem[{Pearl(2012)}]{pearl2012mediation}
Pearl, J. 2012.
\newblock \emph{The mediation formula: A guide to the assessment of causal
  pathways in nonlinear models}.
\newblock Wiley Online Library.

\bibitem[{Pearl and Mackenzie(2018)}]{pearl2018book}
Pearl, J.; and Mackenzie, D. 2018.
\newblock \emph{The book of why: the new science of cause and effect}.
\newblock Basic Books.

\bibitem[{Pearl et~al.(2009)}]{pearl2009causal}
Pearl, J.; et~al. 2009.
\newblock Causal inference in statistics: An overview.
\newblock \emph{Statistics surveys}, 3: 96--146.

\bibitem[{Peters, Janzing, and Sch{\"o}lkopf(2017)}]{PetJanSch17}
Peters, J.; Janzing, D.; and Sch{\"o}lkopf, B. 2017.
\newblock \emph{Elements of Causal Inference - Foundations and Learning
  Algorithms}.
\newblock Adaptive Computation and Machine Learning Series. Cambridge, MA, USA:
  The MIT Press.

\bibitem[{Russell(1913)}]{russell1913notion}
Russell, B. 1913.
\newblock \emph{On the Notion of Cause', reprinted in Mysticism and Logic and
  Other Essays}.
\newblock George Allen \& Unwin.

\bibitem[{Sch{\"o}lkopf(2019)}]{scholkopf2019causality}
Sch{\"o}lkopf, B. 2019.
\newblock Causality for machine learning.
\newblock \emph{arXiv preprint arXiv:1911.10500}.

\bibitem[{Shrestha and Mahmood(2019)}]{shrestha2019review}
Shrestha, A.; and Mahmood, A. 2019.
\newblock Review of deep learning algorithms and architectures.
\newblock \emph{IEEE Access}, 7: 53040--53065.

\bibitem[{Spirtes et~al.(2000)Spirtes, Glymour, Scheines, and
  Heckerman}]{spirtes2000causation}
Spirtes, P.; Glymour, C.~N.; Scheines, R.; and Heckerman, D. 2000.
\newblock \emph{Causation, prediction, and search}.
\newblock MIT press.

\bibitem[{Tenenbaum et~al.(2011)Tenenbaum, Kemp, Griffiths, and
  Goodman}]{Tenenbaum1279}
Tenenbaum, J.~B.; Kemp, C.; Griffiths, T.~L.; and Goodman, N.~D. 2011.
\newblock How to Grow a Mind: Statistics, Structure, and Abstraction.
\newblock \emph{Science}, 331(6022): 1279--1285.

\bibitem[{Wang, Yang, and Wang(2020)}]{Wang2020ProvablyEC}
Wang, L.; Yang, Z.; and Wang, Z. 2020.
\newblock Provably Efficient Causal Reinforcement Learning with Confounded
  Observational Data.
\newblock \emph{ArXiv}, abs/2006.12311.

\bibitem[{Zhang et~al.(2019)Zhang, Lipton, Pineda, Azizzadenesheli, Anandkumar,
  Itti, Pineau, and Furlanello}]{zhang2019learning}
Zhang, A.; Lipton, Z.~C.; Pineda, L.; Azizzadenesheli, K.; Anandkumar, A.;
  Itti, L.; Pineau, J.; and Furlanello, T. 2019.
\newblock Learning causal state representations of partially observable
  environments.
\newblock \emph{arXiv preprint arXiv:1906.10437}.

\bibitem[{Zhang and Bareinboim(2018)}]{zhang2018fairness}
Zhang, J.; and Bareinboim, E. 2018.
\newblock Fairness in decision-making—the causal explanation formula.
\newblock In \emph{Proceedings of the... AAAI Conference on Artificial
  Intelligence}.

\bibitem[{Zhang, Gong, and Sch{\"o}lkopf(2015)}]{zhang2015multi}
Zhang, K.; Gong, M.; and Sch{\"o}lkopf, B. 2015.
\newblock Multi-Source Domain Adaptation: A Causal View.
\newblock In \emph{AAAI}, volume~1, 3150--3157.

\end{thebibliography}
\end{document}